\documentclass{article}

\usepackage{microtype}
\usepackage{graphicx}
\usepackage{subfig}
\usepackage{booktabs} 

\usepackage{setspace}
\usepackage{hyperref}

\DeclareFontEncoding{LS1}{}{}
\DeclareFontSubstitution{LS1}{stix}{m}{n}
\DeclareSymbolFont{mysymbols}       {LS1}{stixscr}  {m} {n}
\DeclareMathSymbol{\squareneswfill} {\mathord}{mysymbols}{"BF}
\DeclareMathSymbol{\squarecrossfill} {\mathord}{mysymbols}{"C0}

\usepackage{svg}
\usepackage{mathtools}
\usepackage{stmaryrd}
\usepackage{blkarray}
\usepackage{mathdots}
\usepackage{multirow}
\usepackage{numprint}
\usepackage{siunitx}
\usepackage{xcolor, colortbl}
\usepackage{tikz}
\usepackage{wrapfig}
\usepackage{enumitem}
\usepackage{soul}
\usepackage{xurl}

\definecolor{convnext}{RGB}{0,114,178}
\definecolor{slak}{RGB}{204,121,167}
\definecolor{dcls}{RGB}{0,158,115}
\definecolor{slak_sparse}{RGB}{255,193,7}
\definecolor{replknet}{RGB}{213,94,0}
\definecolor{lightcream}{RGB}{255,244,212}
\sethlcolor{lightcream}

\definecolor{codegreen}{rgb}{0,0.6,0}
\definecolor{codegray}{rgb}{0.5,0.5,0.5}
\definecolor{codepurple}{rgb}{0.58,0,0.82}
\definecolor{backcolour}{rgb}{0.95,0.95,0.92}

\usepackage{listings}
\lstdefinestyle{mystyle}{
    backgroundcolor=\color{backcolour},   
    commentstyle=\color{codegreen},
    keywordstyle=\color{magenta},
    numberstyle=\tiny\color{codegray},
    stringstyle=\color{codepurple},
    basicstyle=\ttfamily\footnotesize,
    breakatwhitespace=false,         
    breaklines=true,                 
    captionpos=b,                    
    keepspaces=true,                 
    numbers=left,                    
    numbersep=5pt,                  
    showspaces=false,                
    showstringspaces=false,
    showtabs=false,                  
    tabsize=2
}

\DeclareUnicodeCharacter{2212}{-}

\DeclarePairedDelimiter\floor{\lfloor}{\rfloor}

\usepackage[accepted]{icml2023-diffxyz}

\usepackage{amsmath}
\usepackage{amssymb}
\usepackage{mathtools}
\usepackage{amsthm}

\usepackage[capitalize,noabbrev]{cleveref}

\theoremstyle{plain}

\theoremstyle{definition}

\theoremstyle{remark}

\usepackage[textsize=tiny]{todonotes}

\icmltitlerunning{Dilated Convolution with Learnable Spacings: beyond bilinear interpolation.}

\begin{document}

\twocolumn[
\icmltitle{Dilated Convolution with Learnable Spacings: beyond bilinear interpolation}



\icmlsetsymbol{equal}{*}

\begin{icmlauthorlist}
\icmlauthor{Ismail Khalfaoui-Hassani}{aniti,cerco}
\icmlauthor{Thomas Pellegrini}{aniti,irit}
\icmlauthor{Timothée Masquelier}{cerco}
\end{icmlauthorlist}

\icmlaffiliation{aniti}{Artificial and Natural Intelligence Toulouse Institute (ANITI)}
\icmlaffiliation{irit}{IRIT, CNRS, Toulouse INP, Université Toulouse III, Toulouse, France}
\icmlaffiliation{cerco}{CerCo UMR 5549, CNRS, Université Toulouse III, Toulouse, France}

\icmlcorrespondingauthor{Ismail Khalfaoui-Hassani}{ismail.khalfaoui-hassani@univ-tlse3.fr}

\icmlkeywords{deep learning, machine learning, convolution, interpolation, differentiable, Machine Learning, ICML}

\vskip 0.3in
]



\printAffiliationsAndNotice{}  

\begin{abstract}
Dilated Convolution with Learnable Spacings (DCLS) is a recently proposed variation of the dilated convolution in which the spacings between the non-zero elements in the kernel, or equivalently their positions, are learnable. Non-integer positions are handled via interpolation. Thanks to this trick, positions have well-defined gradients. The original DCLS used bilinear interpolation, and thus only considered the four nearest pixels. Yet here we show that longer range interpolations, and in particular a Gaussian interpolation, allow improving performance on ImageNet1k classification on two state-of-the-art convolutional architectures (ConvNeXt and Conv\-Former), without increasing the number of parameters. The method code is based on PyTorch and is available at \href{https://github.com/K-H-Ismail/Dilated-Convolution-with-Learnable-Spacings-PyTorch}{github.com/K-H-Ismail/Dilated-Convolution-with-Learnable-Spacings-PyTorch}.
\end{abstract}

\section{Introduction}
Dilated Convolution with Learnable Spacings (DCLS) is an innovative convolutional method whose effectiveness in computer vision was recently demonstrated~\cite{hassani2023dilated}. In DCLS, the positions of the non-zero elements within the convolutional kernels are learned in a gradient-based manner. The challenge of non-differentiability caused by the integer nature of the positions is addressed through the application of \textbf{bilinear} interpolation. By doing so, DCLS enables the construction of a differentiable convolutional kernel.

DCLS is a differentiable method that only constructs the convolutional kernel. To implement the whole convolution, one can utilize either the native convolution provided by PyTorch or a more efficient implementation such as the ``depthwise implicit gemm" convolution method proposed by \citet{ding2022scaling}, which is suitable for large kernels.

\begin{figure*}[bt]
     \begin{center}
     \subfloat[][]{\includegraphics[width=0.18\textwidth]{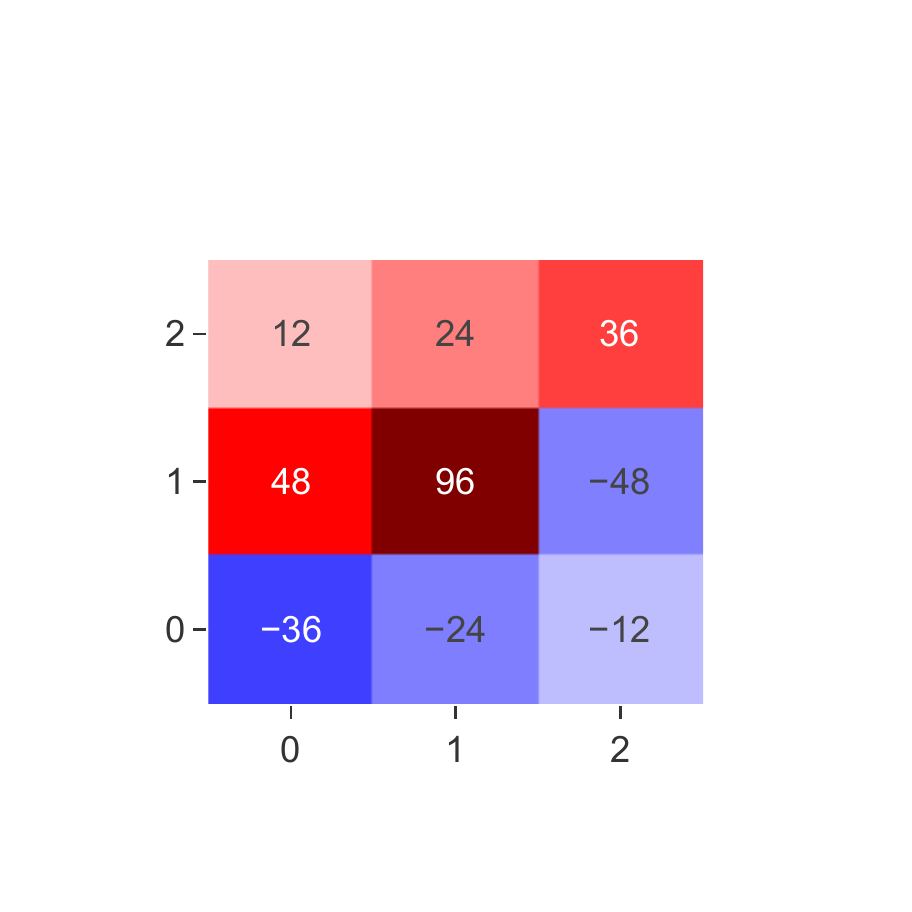}}\hfill
     \subfloat[][]{\includegraphics[width=0.27\textwidth]{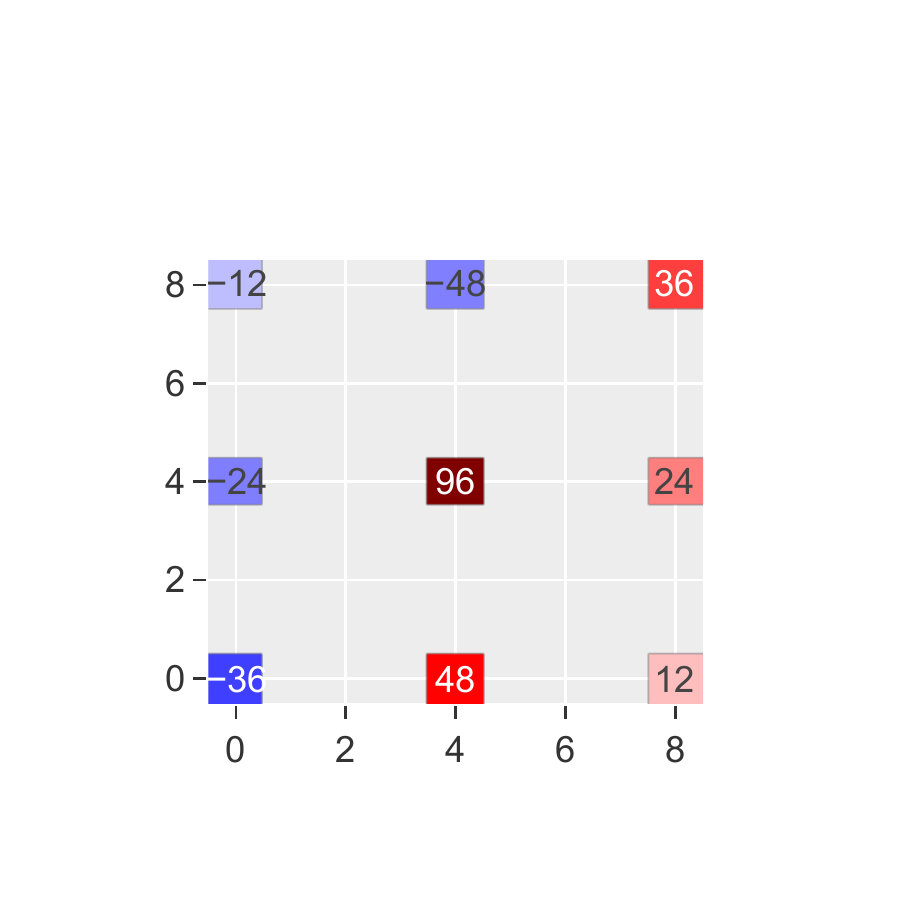}}\hfill
     \subfloat[][]{\label{fig:indexc} \includegraphics[width=0.27\textwidth]{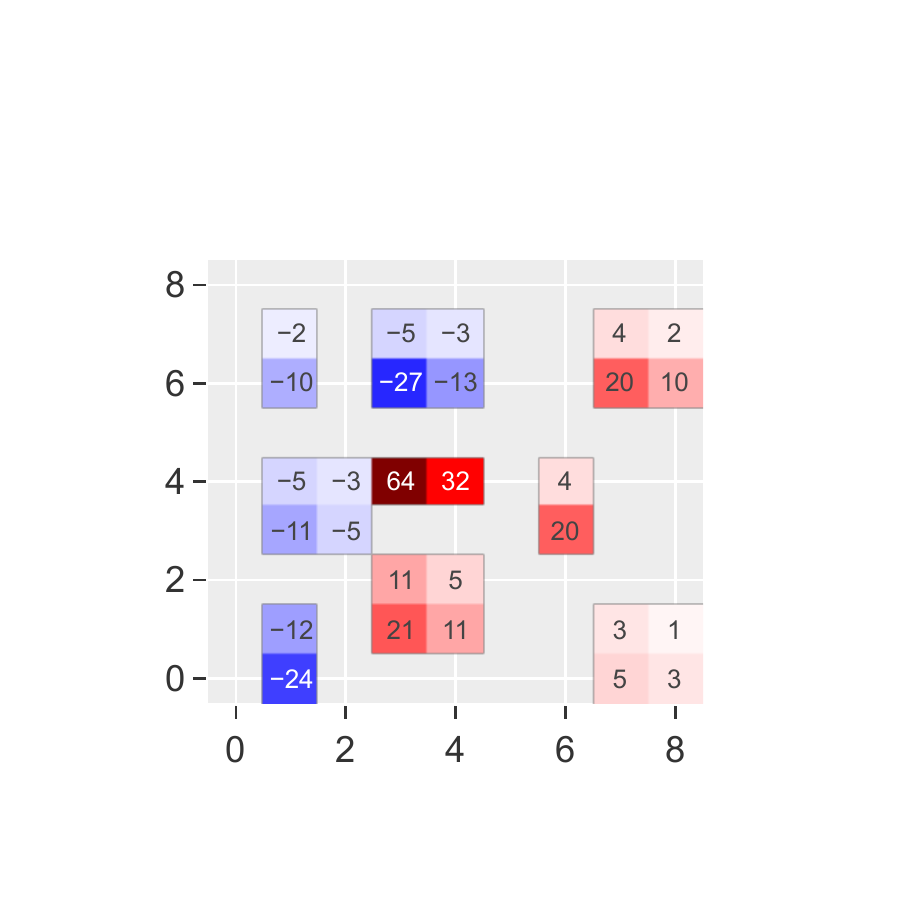}}\hfill     
     \subfloat[][]{\includegraphics[width=0.27\textwidth]{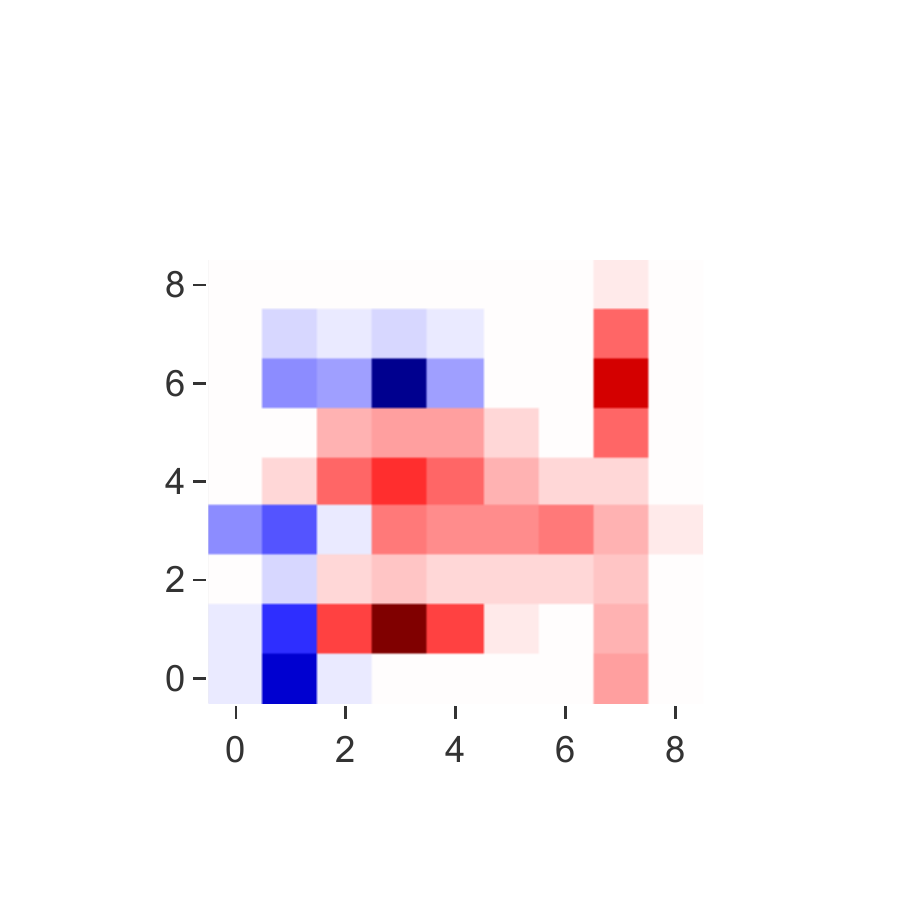}}
    
     \end{center}
    \caption{(a) a standard $3\times 3$ kernel. (b) a standard dilated $3\times 3$ kernel. (c) a 2D-DCLS kernel using bilinear interpolation with 9 kernel elements and a kernel size of 9. (d) the same kernel as (c) with Gaussian interpolation. The numbers have been rounded in all figures and omitted in (d) for readability.}
    \label{fig:index}    
\end{figure*}

The primary motivation behind the development of DCLS was to investigate the potential for enhancing the fixed grid structure imposed by standard dilated convolution in an input-independent way. By allowing an arbitrary number of kernel elements, DCLS introduces a free tunable hyper-parameter called the ``kernel count". Additionally, the ``dilated kernel size" refers to the maximum extent to which the kernel elements are permitted to move within the dilated kernel (Fig.~\ref{fig:indexc}). Both of these parameters can be adjusted to optimize the performance of DCLS. The positions of the kernel elements in DCLS are initially randomized and subsequently allowed to evolve within the limits of the dilated kernel size during the learning process.
The main focus of this paper will be to question the choice of \textbf{bilinear} interpolation used by default in DCLS. We tested several interpolations and found in particular that a \textbf{Gaussian} interpolation with learnable standard deviations made the approach more effective.


To evaluate the effectiveness of DCLS with Gaussian interpolation, we integrate it as a drop-in replacement for the standard depthwise separable convolution in two state-of-the-art convolutional models: the ConvNext-T model~\cite{liu2022convnet} and the ConvFormer-S18 model~\cite{yu2022metaformer}. In Section~\ref{sec:results}, we evaluate the training loss and the classification accuracy of these models on the ImageNet1k dataset \cite{deng2009imagenet}. The remainder of this paper will present a detailed analysis of the methods, equations, algorithms and techniques regarding the application of the Gaussian interpolation in DCLS.

\section{Related work}
In the field of convolutional neural networks (CNNs), various approaches have been explored to improve the performance and efficiency of convolutional operations. Gaussian mixture convolutional networks have investigated the fit of input channels with Gaussian mixtures \cite{celarekgaussian}, while \citet{chen2023gaussian} utilized Gaussian masks in their work. Additionally, continuous kernel convolution was studied in the context of image processing by \citet{kim2023smpconv}. Their approach is similar to the linear correlation introduced in \citet{thomas2019KPConv}. The interpolation function used in the last two works corresponds to the DCLS-Triangle method described in \ref{sec:methods}. Romero et al. have also made notable contributions in learning continuous functions that map the positions to the weights~\cite{romero2022flexconv, romero2022ckconv}.

In the work by~\citet{jacobsen2016structured}, the kernel is represented as a weighted sum of basis functions, including centered Gaussian filters and their derivatives. \citet{pintea2021resolution} extended this approach by incorporating the learning of Gaussian width, effectively optimizing the resolution. \citet{shelhamer2019blurring} introduced a kernel factorization method where the kernel is expressed as a composition of a standard kernel and a structured Gaussian one. In these last three works the Gaussians are centered on the kernel.

Furthermore, the utilization of bilinear interpolation within deformable convolution modules has already shown its effectiveness. \citet{dai2017deformable}, \citet{qi2017deformable} and recently \citet{wang2022internimage} leveraged bilinear interpolation to smoothen the non-differentiable regular-grid offsets in the deformable convolution method. Even more recently, in \citet{kim2023understanding}, a Gaussian attention bias with learnable standard deviations has been successfully used in the positional embedding of the attention module of the ViT model \cite{dosovitskiyimage} and leads to reasonable gains on ImageNet1k.



\section{Methods}
\subsection{From bilinear to Gaussian interpolation }
\label{sec:methods} We denote by $m \in \mathbb{N}^{*}$ the number of kernel elements inside the dilated constructed kernel and we refer to it as the “kernel count”. Moreover, we denote respectively by $s_x, s_y \in \mathbb{N}^{*} \times \mathbb{N}^{* } $, the sizes of the constructed kernel along the x-axis and the y-axis. The latter could be seen as the limits of the dilated kernel, and we refer to them as the “dilated kernel size”.

The $s_x \times s_y$ matrix space over $\mathbb{R}$ is defined as the set of all $s_x \times s_y$ matrices over $\mathbb{R}$, and is denoted $\mathcal{M}_{s_x , s_y}(\mathbb{R})$. 
The real numbers $w$, $p^x$, $\sigma^x$, $p^y$  and $\sigma^y$ respectively stand for the weight, the mean position and standard deviation of that weight along the x-axis (width) and its mean position and standard deviation along the y-axis (height).

The mathematical construction of the 2D-DCLS kernel in \citet{hassani2023dilated} relies on bilinear interpolation and is described as follows :
\begin{align}
\label{eq:2-3}
  \begin{split}
  f \colon \mathbb{R} \times \mathbb{R} \times \mathbb{R} &\to \mathcal{M}_{s_x , s_y } (\mathbb{R})\\
  w, p^x, p^y  & \mapsto \quad K
  \end{split}
\end{align}
where $\forall i\in \llbracket 1 \ .. \ s_x \rrbracket$, $\forall j\in \llbracket 1 \ .. \ s_y \rrbracket  \colon $\\
\begin{equation}
\label{eq:2-4}
\arraycolsep=1.3pt\def\arraystretch{1}
\displaystyle K_{ij} =\left\{\begin{array}{cl}
w \ (1 - r^x)\ (1 - r^y) & \text {if } i = \floor{p^x}, \ j = \floor{p^y} \\
w \ r^x \ (1 - r^y) & \text {if }  i = \floor{p^x} + 1, \ j = \floor{p^y} \\
w \ (1 - r^x) \ r^y & \text {if }  i = \floor{p^x}, \ j = \floor{p^y} + 1\\
w \ r^x \ r^y  & \text {if }  i = \floor{p^x} {+} 1, \ j = \floor{p^y} {+} 1 \\
0 & \text {otherwise } 
\end{array}\right.
\end{equation}
and where the fractional parts are:
\begin{equation}
    \begin{array}{ccc}
    r^x = \{p^x\} = p^x - \floor{p^x} & \text{and} & r^y = \{p^y\} = p^y - \floor{p^y}
    \end{array}
\end{equation}

An equivalent way of describing the constructed kernel $K$ in Equation~\ref{eq:2-4} is:
\begin{equation}
    \label{eq:4}
    K_{ij} = w \cdot g(p^x - i) \cdot g(p^y - j) 
\end{equation}
with 
\begin{equation}
    g \colon x \mapsto \text{max}(0, \ 1 - |x|)
\end{equation}
This expression corresponds to the bilinear interpolation as described in \citet[][eq. 4]{dai2017deformable}.

\begin{figure}[bt]
     \centering
     \includegraphics[width=0.275\textwidth]{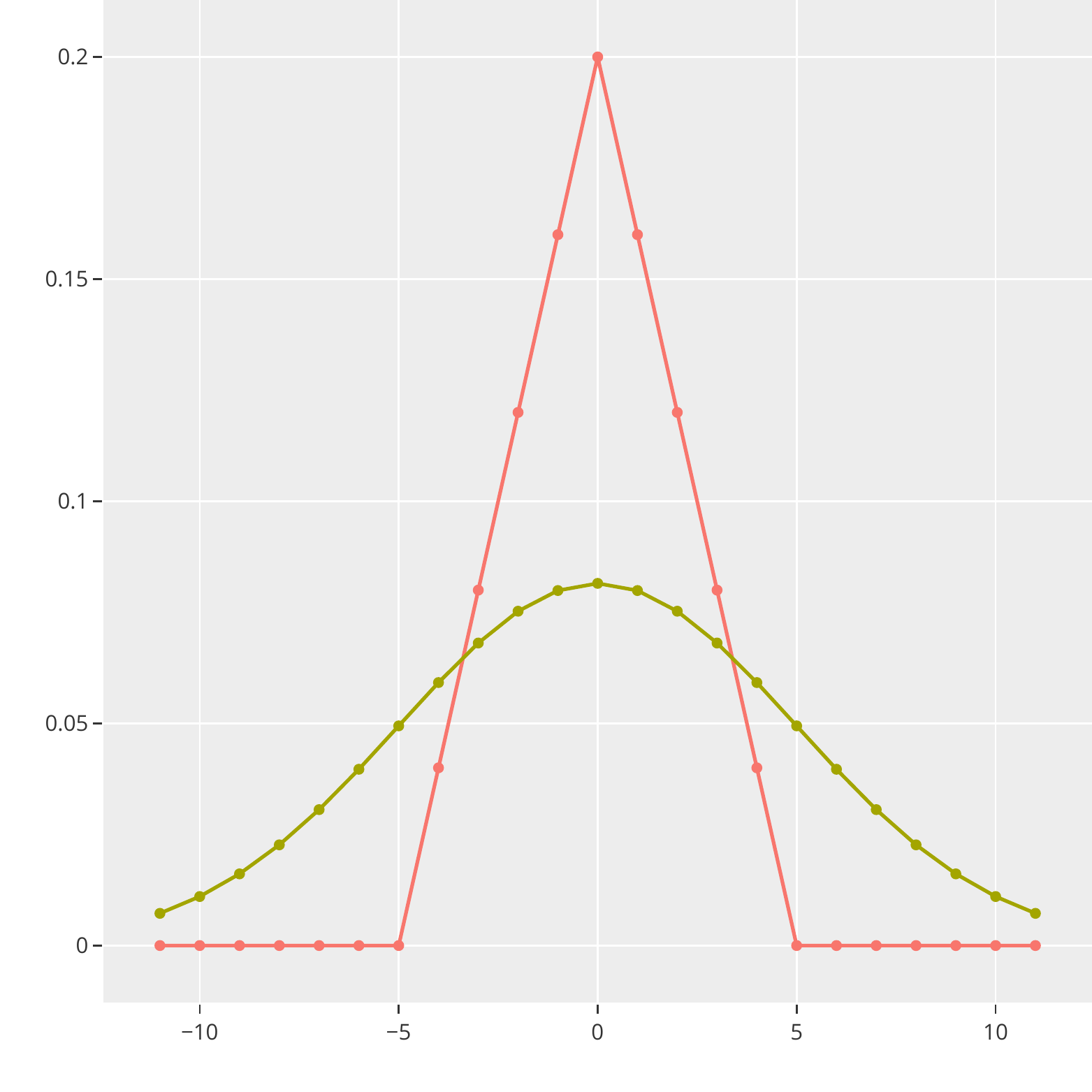}
     \caption{1D view of Gaussian and $\Lambda$ functions with $\sigma = 5$.}
     \label{weightings}
\end{figure}

In fact, this last $g$ function is known as the triangle function (refer to Fig.~\ref{weightings} for a graphic representation), and is widely used in kernel density estimation. From now on, we will note it as
\begin{equation}
    \forall x \in \mathbb{R} \quad \quad \Lambda (x) \overset{\text{def}}{=} \text{max}(0, \ 1 - |x|)
\end{equation}

First, we consider a scaling by a parameter $\sigma \in \mathbb{R}_+$ for the triangle function (the bilinear interpolation corresponds to $\sigma=1$),
\begin{equation}
\label{eq:triangle}
    \forall x \in \mathbb{R}, \quad \forall \sigma \in \mathbb{R}_+ \quad \Lambda_\sigma (x) \overset{\text{def}}{=} \text{max}(0, \ \sigma - |x|)
\end{equation}
We found that this scaling parameter $\sigma$ could be learned by backpropagation and that doing so increases the performance of the DCLS method. As we have different $\sigma$ parameters for the x and y-axes in 2D-DCLS, learning the standard deviations costs two additional learnable parameters and two additional FLOPs (multiplied by the number of the channels of the kernel and the kernel count). We refer to the DCLS method with triangle function interpolation as the DCLS-Triangle method.

Second, we tried a smoother function rather than the piecewise affine triangle function, namely the Gaussian function:
\begin{equation}
\label{eq:gauss}
    \begin{array}{lcrr}
     \forall x \in \mathbb{R}, \ \forall \sigma \in \mathbb{R}^*, & G_\sigma (x) \overset{\text{def}}{=} \text{exp}\left({-  {\dfrac{x^2}{2\sigma^2}}}\right) & &
    \end{array}
\end{equation}
We refer to the DCLS method with Gaussian interpolation as the DCLS-Gauss method. In practice, instead of Equations \ref{eq:triangle} and \ref{eq:gauss}, we respectively use:
\begin{equation}
    \forall x \in \mathbb{R}, \ \forall \sigma \in \mathbb{R}, \enskip \Lambda_{\sigma_0 + \sigma} (x) = \text{max}(0, \ \sigma_0 + |\sigma| - |x|)
\end{equation}
\begin{equation}
     \forall x \in \mathbb{R}, \ \forall \sigma \in \mathbb{R}, \enskip G_{\sigma_0 + \sigma} (x) = \text{exp}\left({-  \dfrac{1}{2} \dfrac{x^2}{(\sigma_0 + |\sigma|)^2}}\right)
\end{equation}
with $\sigma_0 \in \mathbb{R}^*_+$ a constant that determines the minimum standard deviation that the interpolation could reach. For the triangle interpolation, we take $\sigma_0 = 1$ in order to have at least 4 adjacent interpolation values (see Figure~\ref{fig:indexc}). And for the Gaussian interpolation, we set $\sigma_0 = 0.27$. 

Last, to make the sum of the interpolation over the dilated kernel size equal to 1, we divide the interpolations by the following normalization term :
\begin{equation} 
A = \epsilon + \sum_{i=1}^{s_x}\sum_{j=1}^{s_y} \mathcal{I}_{\sigma_0 + \sigma^x}(p^x -i) \cdot \mathcal{I}_{\sigma_0 + \sigma^y}(p^y -j) 
\end{equation} with $\mathcal{I}$ an interpolation function ($\Lambda$ or $G$ in our case) and $\epsilon = 1e-7$ for example, to avoid division by zero.

\textbf{Other interpolations} Based on our tests, other functions such as Lorentz, hyper-Gaussians and sinc functions have been tested with no great success. In addition, learning a correlation parameter $\rho \in [-1,1]$ or equivalently a rotation parameter $\theta \in [0, 2\pi]$ as in the bivariate normal distribution density did not improve performance (maybe because cardinal orientations predominate in natural images).
\subsection{The 2D-DCLS-Gauss kernel construction algorithm}
\label{sec:algo}
In the following, we describe with pseudocode the  kernel construction used in 2D-DCLS-Gauss and 2D-DCLS-Triangle. $\mathcal{I}$ is the interpolation function ($\Lambda$ or $G$ in our case) and $\epsilon = 1e-7$. In practice, $w$, $p^x$, $p^y$, $\sigma^x$  and $\sigma^y$ are 3-D tensors of size (\texttt{channels\_out, channels\_in // groups, K\_count}), but the algorithm presented here is easily extended to this case by applying it channel-wise.
\begin{algorithm}[ht]
    \setstretch{1.2}
    \begin{algorithmic}[1]

        \REQUIRE $w$, $p^x$, $p^y$, $\sigma^x$, $\sigma^y$ :  vectors of dimension $m$
        \ENSURE $K$ : the constructed kernel, of size ($s_x \times s_y$)
        \STATE $K \leftarrow 0_{s_x,s_y}$ \COMMENT{zero tensor of size $s_x, s_y$} 
        \label{alg:forward}        
        \FOR{$k=0$ {\bfseries to} $m-1$}
            \STATE $H \leftarrow 0_{s_x,s_y}$
            \STATE $p_k^x \leftarrow p_k^x + s_x // 2 $; $\quad p_k^y \leftarrow p_k^y + s_y // 2 $ 
            \STATE $\sigma_k^x \leftarrow |\sigma_k^x| + \sigma_0^{\mathcal{I}}$;   $\quad \sigma_k^y \leftarrow|\sigma_k^y| + \sigma_0^{\mathcal{I}}$ 
            
            \FOR{$i=0$ {\bfseries to} $s^x-1 $}
                \FOR{$j=0$ {\bfseries to} $s^y-1 $}
                \STATE $H[i,j] \leftarrow \mathcal{I}_{\sigma_k^x}(p_k^x - i) * \mathcal{I}_{\sigma_k^y}(p_k^y - j)$
                \ENDFOR 
            \ENDFOR        
        \STATE $H[:,:] \leftarrow H[:,:] \ / (\epsilon + \sum\limits_{i=0}^{s^x-1} \sum\limits_{j=0}^{s^y-1} H[i,j])$
        \STATE $K \leftarrow K + H * w_k$      
        \ENDFOR
    \end{algorithmic} 
    \caption{2D-DCLS-interpolation kernel construction}
    \label{algo:1}
\end{algorithm}
\begin{table*}[!bt]
\caption{\textbf{Classification accuracy on the validation set and training loss on ImageNet-1K.} For the 17/34 bilinear, the 23/26 Triangle and Gaussian cases, the results have been averaged over 3 distinct seeds (the corresponding lines are highlighted in yellow).}
\begin{center}
\begin{small}
\begin{sc}
$$
\begin{array}{lcccccc}
\toprule
\text { model @ 224} & \begin{array}{l}
\text { ker. size } \\
\text { / count  }
\end{array} & \text {interpolation}  & \text { \# param.} & \text { train loss } & \text { Top-5 acc.} & \text { Top-1 acc.} \\
\hline 
\abovespace
\text { ConvNeXt-T }\squareneswfill  & 7^2 \ / \ 49 &  & 28.59 \mathrm{M} & 2.828 & 96.05 & 82.08 \\
\rowcolor{lightcream}\text { ConvNeXt-T } \squarecrossfill & 17^2  \ / \  34 & \text{Bilinear} &  28.59 \mathrm{M} & 2,775 & 96.11 & 82.44 \\
\rowcolor{lightcream}\text { ConvNeXt-T } \odot & 23^2  \ / \  26 & \text{Triangle} &  28.59 \mathrm{M} & 2.787  & 96.09 & 82.34 \\
\rowcolor{lightcream}\text { ConvNeXt-T } \star & 23^2  \ / \  26 & \text{Gaussian} & 28.59 \mathrm{M} & 2.762 & 96.18 & 82.44 \\
\text { ConvNeXt-T } & 17^2  \ / \  26 & \text{Gaussian} &  28.59 \mathrm{M} & 2.773 & 96.17 & 82.40 \\
\text { ConvNeXt-T } & 23^2  \ / \  34 & \text{Gaussian} & 28.69 \mathrm{M} & 2.758 & 96.22 & 82.60
\\
\midrule
\text { ConvFormer-S18 } \squareneswfill &  7^2 \ / \ 49 & & 26.77 \mathrm{M} & 2.807 & 96.17 & 82.84 \\
\rowcolor{lightcream}\text { ConvFormer-S18 } \squarecrossfill & 17^2 \ / \ 40 &  \text{Bilinear}  & 26.76 \mathrm{M} & 2.764 & 96.42 & 83.14\\
\rowcolor{lightcream}\text { ConvFormer-S18 } \odot & 23^2 \ / \ 26 &  \text{Triangle}  & 26.76 \mathrm{M} & 2.761 & 96.38 & 83.09\\
\rowcolor{lightcream}\text { ConvFormer-S18 }  \star & 23^2 \ / \ 26 &  \text{Gaussian} & 26.76 \mathrm{M} & 2.747 & 96.31 & 82.99\\
\bottomrule
\end{array}
$$
\end{sc}
\end{small}
\end{center}
\label{tab:results}
\end{table*}
\section{Learning techniques}
\label{sec:usage}
Having discussed the implementation of the interpolation in the DCLS method, we now shift our focus to the techniques employed to maximize its potential. We retained most of the techniques used in \citet{hassani2023dilated}, and suggest new ones for learning standard deviations parameters. In Appendix~\ref{app:learning_techniques}, we present the training techniques that have been selected based on consistent empirical evidence, yielding improved training loss and validation accuracy.
\section{Results}
\label{sec:results}

\begin{figure}[bt]
     \centering
     \includegraphics[width=0.49\textwidth]{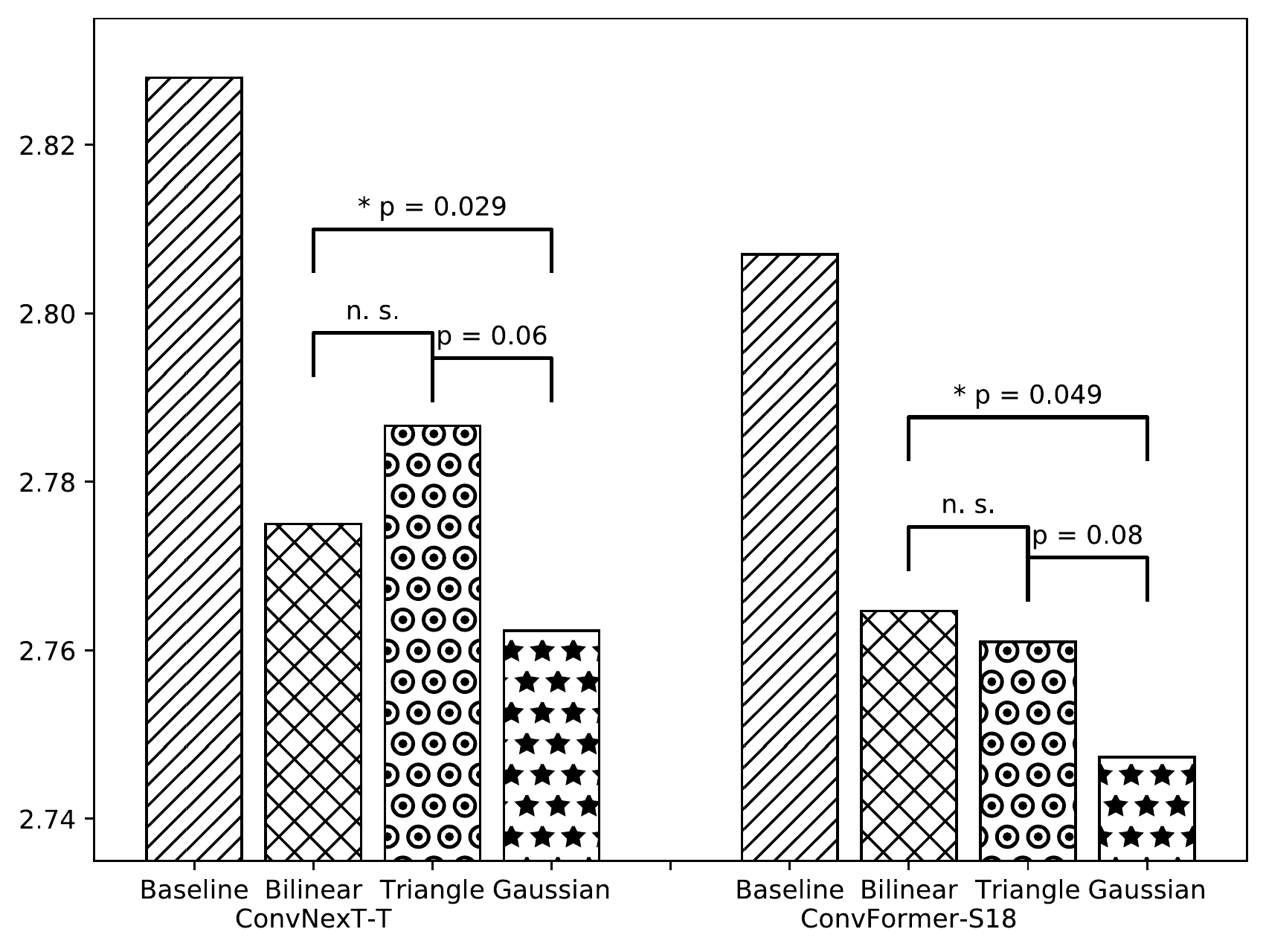}
     \caption{Training loss for ConvNeXt-T and ConvFormer-S18 models with DCLS according to interpolation type (lower is better). The pairwise p-values have been calculated using an independent two-sample Student t-test assuming equal variances. The vertical line segments stand for the standard errors. }
     \label{t-test}
\end{figure}
We took two recent state-of-the-art convolutional architectures, ConvNeXt and ConvFormer, and drop-in replaced all the depthwise convolutions by DCLS ones, using the three different interpolations (bilinear, triangle or Gauss). Table~\ref{tab:results} reports the results in terms of training loss and validation accuracy.

A first observation is that all the DCLS models perform much better than the baselines, whereas they have the same number of parameters. There are also subtle differences between interpolation functions. As Figure~\ref{t-test} shows, triangle and bilinear interpolations perform similarly, but the Gaussian interpolation performs significantly better. 

Furthermore, the advantage of the Gaussian interpolation w.r.t. bilinear is not only due to the use of a larger kernel, as a 17x17 Gaussian kernel (5th line in Table~\ref{tab:results}) still outperforms the bilinear case (2nd line). Finally, the 6th line in Table~\ref{tab:results} shows that there is still room for improvement by increasing the kernel count, although this slightly increases the number of trainable parameters w.r.t. the baseline.

\section{Conclusion}
In conclusion, this study introduces Gaussian and $\Lambda$ interpolation methods as alternatives to bilinear interpolation in Dilated Convolution with Learnable Spacings (DCLS). Evaluations on state-of-the-art convolutional architectures demonstrate that Gaussian interpolation improves performance of image classification task on ImageNet1k without increasing parameters. Future work could implement the Whittaker-Shannon interpolation instead of the Gaussian interpolation and search for a dedicated architecture, that will make the most of DCLS.
\section*{Acknowledgments}
This work was performed using HPC resources from GENCI–IDRIS (Grant 2021-[AD011013219]). Support from the ANR-3IA Artificial and Natural Intelligence Toulouse Institute is gratefully acknowledged. We would also like to thank the region of Toulouse Occitanie.

\bibliography{main}
\bibliographystyle{icml2023}

\newpage
\appendix
\onecolumn
\section{Code and reproducibility}
The code of the method is based on PyTorch and available at \href{https://github.com/K-H-Ismail/Dilated-Convolution-with-Learnable-Spacings-PyTorch}{https://github.com/K-H-Ismail/Dilated-Convolution-with-Learnable-Spacings-PyTorch}.

\section{\emph{Pytorch} implementation of the 2D-DCLS-Gauss and 2D-DCLS-Triangle forward algorithm}
\label{code:pytorch}
\begin{lstlisting}[language=Python]
class ConstructKernel2d(Module):
    def __init__(self, out_channels, in_channels, groups, kernel_count, dilated_kernel_size, version):
        super().__init__()
        self.version = version
        self.out_channels, self.in_channels = out_channels, in_channels
        self.groups = groups
        self.dilated_kernel_size = dilated_kernel_size
        self.kernel_count = kernel_count
        self.IDX, self.lim = None, None

    def __init_tmp_variables__(self, device):
        if self.IDX is None or self.lim is None:
            J = Parameter(torch.arange(0, self.dilated_kernel_size[0]), 
            requires_grad=False).to(device)
            I = Parameter(torch.arange(0, self.dilated_kernel_size[1]),
            requires_grad=False).to(device)
            I = I.expand(self.dilated_kernel_size[0],-1)
            J = J.expand(self.dilated_kernel_size[1],-1).t()
            IDX = torch.cat((I.unsqueeze(0),J.unsqueeze(0)), 0)
            IDX = IDX.expand(self.out_channels, self.in_channels//self.groups, 
            self.kernel_count,-1,-1,-1).permute(4,5,3,0,1,2)
            self.IDX = IDX
            lim = torch.tensor(self.dilated_kernel_size).to(device)
            self.lim = lim.expand(self.out_channels,
            self.in_channels//self.groups, self.kernel_count, -1).permute(3,0,1,2)
        else:
            pass

    def forward_vtriangle(self, W, P, SIG):
        P = P + self.lim // 2
        SIG = SIG.abs() + 1.0
        X = (self.IDX - P)
        X = ((SIG - X.abs()).relu()).prod(2)
        X  = X / (X.sum((0,1)) + 1e-7) # normalization
        K = (X * W).sum(-1)
        K = K.permute(2,3,0,1)
        return K

    def forward_vgauss(self, W, P, SIG):
        P = P + self.lim // 2
        SIG = SIG.abs() + 0.27
        X = ((self.IDX - P) / SIG).norm(2, dim=2)
        X = (-0.5 * X**2).exp()
        X  = X / (X.sum((0,1)) + 1e-7) # normalization
        K = (X * W).sum(-1)
        K = K.permute(2,3,0,1)
        return K
 
    def forward(self, W, P, SIG):
        self.__init_tmp_variables__(W.device)
        if self.version == 'triangle':
            return self.forward_vtriangle(W, P, SIG)
        elif self.version == 'gauss':
            return self.forward_vgauss(W, P, SIG)        
        else:
            raise
\end{lstlisting}

\section{Learning techniques}
\label{app:learning_techniques}
\begin{itemize}[leftmargin=*]
    \item \textbf{Weight decay:} No weight decay was used for positions. We apply the same for standard deviation parameters. 
    \item \textbf{Positions and standard deviations initialization:} position parameters were initialized following a centered normal law of standard deviation 0.5. Standard deviation parameters were initialized to a constant $0.23$ in DCLS-Gauss and to $0$ in DCLS-Triangle in order to have a similar initialisation to DCLS with bilinear interpolation at the beginning.
    \item \textbf{Positions clamping :} Previously in DCLS, kernel elements that reach the dilated kernel size limit were clamped. It turns out that this operation is no longer necessary with the Gauss and $\Lambda$ interpolations. 
    \item \textbf{Dilated kernel size tuning:} When utilizing bilinear interpolation in ConvNeXt-dcls, a dilated kernel size of 17 was found to be optimal, as larger sizes did not yield improved accuracy. However, with Gaussian and $\Lambda$ interpolations, there appears to be no strict limit to the dilated kernel size. Accuracy tends to increase logarithmically as the size grows, with improvements observed up to kernel sizes of 51. It is important to note that increasing the dilated kernel size does not impact the number of trainable parameters, but it does affect throughput. Therefore, a compromise between accuracy and throughput was achieved by setting the dilated kernel size to 23.
    \item \textbf{Kernel count tuning:} This hyper-parameter has been configured to the maximum integer value while still remaining below the baselines to which we compare ourselves in terms of trainable parameters. It is worth noting that each additional element in the 2D-DCLS-Gauss or 2D-DCLS-Triangle methods introduces five more learnable parameters: weight, vertical and horizontal position, and their respective standard deviations. To maintain simplicity, the same kernel count was applied across all model layers.
    \item \textbf{Learning rate scaling:} To maintain consistency between positions and standard deviations, we applied the same learning rate scaling ratio of 5 to both. In contrast, the learning rate for weights remained unchanged.
    \item \textbf{Synchronizing positions:} we shared the kernel positions and standard deviations across convolution layers with the same number of parameters, without sharing the weights. Parameters in these stages were centralized in common parameters that accumulate the gradients. 
    
\end{itemize}
\section{1D and 3D convolution cases} 
For the 3D case, Equation~\ref{eq:4} can be generalized as a product along spatial dimensions. We denote respectively by $s_x, s_y, s_z \in \mathbb{N}^{*} \times \mathbb{N}^{* } \times \mathbb{N}^{* } $, the sizes of the constructed kernel along the x-axis, the y-axis and the z-axis. The constructed kernel tensor $K^{3D} \in \mathcal{M}_{s_x , s_y, s_z } (\mathbb{R})$ is therefore:

$\forall i\in \llbracket 1 \ .. \ s_x \rrbracket$, $\forall j\in \llbracket 1 \ .. \ s_y \rrbracket$, $\forall k\in \llbracket 1 \ .. \ s_z \rrbracket  \colon $
\begin{equation}
    K^{3D}_{ijk} = w \cdot \mathcal{I}_{\sigma_0 + \sigma^x}(p^x - i) \cdot \mathcal{I}_{\sigma_0 + \sigma^y}(p^y - j) \cdot \mathcal{I}_{\sigma_0 + \sigma^z}(p^z - k) 
\end{equation}
with $\mathcal{I}$ an interpolation function ($\Lambda$ or $G$), $\sigma_0 = 1$ for the $\Lambda$ interpolation and $\sigma_0 = 0.27$ for the Gaussian one. $w$, $p^x$, $\sigma^x$, $p^y$, $\sigma^y$, $p^z$  and $\sigma^z$ respectively representing  the weight, the mean position and standard deviation of that weight along the x-axis (width), the mean position and standard deviation along the y-axis (height) and its mean position and standard deviation along the z-axis (depth).

The constructed kernel vector $K^{1D} \in \mathbb{R}^{s_x}$ for the 1D case is simply:

$\forall i\in \llbracket 1 \ .. \ s_x \rrbracket \colon $
\begin{equation}
    K^{1D}_{i} = w \cdot \mathcal{I}_{\sigma_0 + \sigma^x}(p^x - i) 
\end{equation}

The Algorithm~\ref{algo:1} as well as the Pytorch code~\ref{code:pytorch} are readily adapted to these cases by following the above note.

\end{document}